\documentclass[12pt, titlepage, reqno]{article}

 \usepackage[cmex10]{amsmath}

 \usepackage{amssymb, latexsym}
 \usepackage{amsmath} 
 \usepackage{amsthm}
 \usepackage{bm}
 \usepackage[mathscr]{eucal}
 \usepackage{enumerate}

 \usepackage{natbib}
 \usepackage[french,english]{babel}

\usepackage{algorithm,algpseudocode}
\usepackage{caption}
\usepackage{color}

\usepackage{gensymb}

\usepackage[caption = false]{subfig}

\usepackage{float}

\usepackage{algpseudocode}

\usepackage{graphicx}

\usepackage{tabularx}
\usepackage{slashbox}

\usepackage{hyperref}
\hypersetup{
    colorlinks,%
    citecolor=blue,%
    filecolor=blue,%
    linkcolor=blue,%
    urlcolor=blue,
}  
  
\usepackage{multirow}

\graphicspath{{./Figures/}}

\begin{document}

 \begin{titlepage}

 \begin{center}
\textbf{Investigation on the use of Hidden-Markov Models in automatic transcription of music}

\vspace{10ex}

Dorian Cazau$^{a}$ \footnote{Corresponding author e-mail: dorian.cazau@ensta-bretagne.fr}, Gregory Nuel$^{b}$

\vspace{1ex}

\begin{flushleft}

\small{$^a$ ENSTA Bretagne - Lab-STICC (UMR CNRS 6285), Universit\'e Europ\'eenne de Bretagne, 2 rue Fran\c{c}ois Verny, 29806 Brest Cedex 09, France} \\

\small{$^b$ Laboratoire de Math\'ematiques Appliqu\'ees (MAP5) UMR CNRS 8145, Universit\'e Paris Descartes, Paris, France} 

\end{flushleft}

 \end{center}

 \end{titlepage}

\begin{abstract}

Hidden Markov Models (HMMs) are a ubiquitous tool to model time series data, and have been widely used in two main tasks of Automatic Music Transcription (AMT): note segmentation, i.e. identifying the played notes after a multi-pitch estimation, and sequential post-processing, i.e. correcting note segmentation using training data. In this paper, we employ the multi-pitch estimation method called Probabilistic Latent Component Analysis (PLCA), and develop AMT systems by integrating different HMM-based modules in this framework. For note segmentation, we use two different two-state on/off HMMs, including a higher-order one for duration modeling. For sequential post-processing, we focused on a musicological modeling of polyphonic harmonic transitions, using a first- and second-order HMMs whose states are defined through candidate note mixtures. These different PLCA plus HMM systems have been evaluated comparatively on two different instrument repertoires, namely the piano (using the MAPS database) and the marovany zither. Our results show that the use of HMMs could bring noticeable improvements to transcription results, depending on the instrument repertoire.

\end{abstract}

\section{Introduction}\label{sec:intro}

Work on Automatic Music Transcription (AMT) dates back more than 30 years, and has known numerous applications in the fields of music information retrieval, interactive computer systems, and automated musicological analysis \citep{Klapuri2004b}. Due to the difficulty in producing all the information required for a complete musical score, AMT is commonly defined as the computer-assisted process of analyzing an acoustic musical signal so as to write down the musical parameters of the sounds that occur in it, which are basically the pitch, onset time, and duration of each sound to be played. Despite a large enthusiasm for AMT challenges, and several audio-to-MIDI converters available commercially, perfect polyphonic AMT systems are out of reach of today's technology \citep{Klapuri2004b,Benetos2013c}. 

To overcome these limitations, a practical engineering solution was to use computational techniques from statistics and digital signal processing, allowing more complex modeling of the musical signal. In this paper, we investigate the use of different Hidden Markov Models (HMMs) in AMT, and evaluate their impacts on transcription performance. HMMs are a ubiquitous tool to model time series data, and have been widely used in various tasks of Music Information Retrieval, especially in music structure analysis by characterizing repetitive patterns \citep{Logan2000} or performing harmonic analysis \citep{Raphael2003}, chord estimation \citep{Lee2008} and musicological modeling of note transitions \citep{Ryynanen2008}. For what concerns the task of AMT, the sequential structure that may be inferred from musical signals can be usefully integrated to systems with HMMs. Indeed, as approaches consisting of a multi-pitch estimation stage display the obvious fault of processing each frame independently of its neighbors, without exploiting the inherent temporal structure of music. HMMs have then been used for three main tasks: note acoustic modeling (e.g. \citep{Ryynanen2005}), note segmentation (i.e. identifying the played notes after a multi-pitch estimation) and sequential post-processing (i.e. correcting note segmentation using training data). In this paper, we will review and develop original HMM models for these two last tasks. For note segmentation, we use two different two-state on/off HMMs, including the first-order one developed by \citep{Poliner2007}, and an original higher-order one for duration modeling. For sequential post-processing, we focused on a musicological modeling of polyphonic harmonic transitions, using a first- and second-order HMMs on estimated states made of candidate note mixtures. These different HMMs are integrated to a Probabilistic Latent Component Analysis  \citep{Smaragdis2006} framework, used to perform our multi-pitch estimation. PLCA is a spectrogram factorization method, able to model a magnitude spectrogram as a linear combination of spectral vectors from a dictionary. These different PLCA plus HMMs systems have been evaluated comparatively on two different sound datasets, the classical piano (MAPS database) and the marovany zither from Madagascar.

\section{Methods}\label{sec:meth}

\subsection{Probabilistic Latent Component Analysis (PLCA)}

\subsubsection{Multi-Pitch Estimation}

We present in this section our baseline AMT system, following a classical PLCA model drawn from previous studies \citep{Benetos2013d}. In the Probabilistc Latent Component Analysis (PLCA) is a probabilistic factorization method, based on the assumption that a suitably normalized magnitude spectrogram, V, can be modeled as a joint distribution over time and frequency, $P(f,t)$, with f is the log-frequency index and t the time index. This quantity can be factored into a frame probability $P(t)$, which can be computed directly from the observed data (i.e. energy spectrogram), and a conditional distribution over frequency bins $P(f|t)$, as follows

\begin{equation}\label{EqPLCA}
P(f|t) = P(t) \sum_{i,m} P(f|i,m) P(m|i,t) P(i|t)
\end{equation}

where $P(f|i,m)$ are the spectral templates for pitch $i \in\ I$ (with I the number of pitches) and playing mode m, $P(m|i,t)$ is the playing mode activation, and $P(i|t)$ is the pitch activation (i.e. the transcription). In this paper, the playing mode m will refer to different dynamics of instrument playing (i.e. note loudness). To estimate the model parameters $P(m|i,t)$ and $P(m|t)$, since there is usually no closed-form solution for the maximization of the log-likelihood or the posterior distributions, iterative update rules based on the Expectation-Maximization (EM) algorithm \citep{Dempster1977} are employed. The spectral templates $P(f|i,m)$ are extracted from isolated note samples using a one component PLCA, and are not updated.

\subsubsection{Note segmentation based on a simple thresholding}\label{ThreshdolingNoteDec}

Eventually, as in most spectrogram factorization-based transcription or pitch tracking methods \citep{Grindlay2011,Mysore2009,Dessein2010}, we use a simple threshold-based detection of the note activations from the pitch activity matrix $P(i,t)$, followed by a minimum duration pruning. This threshold for detection was set empirically to 0.02 and the minimum duration for pruning was set to 50 ms. This operation results in a list of note candidates $\tilde{N}(i,k)$ of size $N_c$, with k being the index of the activation interval $T_{i,k}$ for pitch i.

\subsection{Hidden Markov Model (HMM)}

\subsubsection{First-Order HMM}

Consider a system that is described by a set of N distinct states $S_k$, where $S_k$ $\in$ \textbf{S} = $\{S_1,S_2,...,S_N\}$. The states of the system may change with time, and at the time instants t, they are denoted by $q_t$, a discrete-time random variable taking value in the finite set \textbf{S}. In most modeling scenarios, the state sequence is not directly observable, but hidden from the observer. Then Hidden Markov Model (HMM) assumes that each state is associated with an emission probability density function that generates our observed data $y_t$ at every time instant t. A HMM can then be seen as a doubly stochastic generative process, with two components: a set of hidden variables that can not be observed directly from the data, and a Markov property that is usually related to some dynamical temporal behaviour of the hidden variables. 

It is very complex and difficult to directly model the joint probability density function of the observation sequence. A reasonable approach is to group nearby observations of similar characteristics as being produced by the same state and then consider how do the states progress and how does a state sequence produce the observation sequence. Hence, the model is split into two parts, the first part considers the probability of a state sequence and the second part considers the observation sequence based on a state sequence, and the joint probability distribution given by the model is written as, given a sequence of measurements $y_1,..., y_T$ and assuming a certain sequence of hidden states $q_1,..., q_T$,

\vspace{-0.5cm}
\begin{multline}\label{JointProba}
P(q_1,..., q_T,y_1,..., y_T) = P(q_1)P(y_1 | q_1) \\
\prod_{t=2}^T P(y_t | q_t) \cdot P(q_t | q_{t-1})
\end{multline}

and the most likely state sequence Q is given by 

\begin{equation}
Q = arg \max\limits_{q_t} \{ P(q_1,..., q_T,y_1,..., y_T) \}
\end{equation}

which can be estimated by the Viterbi algorithm \citep{Dempster1977}. 

\subsubsection{Higher-Order HMM (HO-HMM)}

So far, we have made the assumption that the future state of a process depends only on its single most recent state, in terms of state transition and output observation. Consequently, the feature vectors of consecutive sound frames belonging to the same state are independent identically distributed, and trajectory modeling (i.e., frame correlation) in the frame space is not included. Such modeling characteristics are often unreasonable and reveal limits of first-order HMMs, especially in regards to state duration modeling as they violate the high correlations among successive frame-wise states. Indeed, the piecewise stationary assumption does not precisely match the non-stationary nature of the actual sound process. 

In regards to such limitations, it sometimes makes sense to take more than one previous state into account using a higher-order Markov chain to get a higher prediction accuracy. HO-HMMs can be defined by making transition probabilities depend on the last N states in addition to the most recent state. HO-HMMs have received a great deal of interest to model frame state duration \citep{Mari1997,Ching2008}. By definition, one can actually build more ``memory" into states by using a high-order Markov model with a trajectory modeling within state sequences.

Mathematically, if we consider that state transition depends only on the previous $O_d$ states and output depends only on the previous $O_{e}$ states, then the model can be thought as an HO-HMM of orders ( $O_d$, $O_{e}$). The joint probability of state and observation sequences of eq. \ref{JointProba} can then be re-written as

\vspace{-0.5cm}
\begin{multline}\label{EqHOHMM}
P(q_1,..., q_T,y_1,..., y_T ) = P(q_1)P(y_1 | q_1) 
\\
\cdot  P(q_2 |q_1)P(y_2 | q_1,q_2) 
\\
\cdot P(q_t | q_{t-O_d},...,q_{t-1}) P(y_t | q_{t-c+1},...,q_{t} ) 
\\
\cdot P(q_T | q_{T-O_d},...,q_{T-1} )P(y_T | q_{T-c+1},...,q_{T}) 
\end{multline}

From eq. \ref{EqHOHMM}, we can see that there is an enormous number of parameters to be estimated (about $N^{O_d}$ transition probabilities and $N^{O_{e}}$ probability density functions for an N-state HO-HMM). The corresponding transition matrix \textbf{A} is an $N^{O_d} \times N$ matrix. Obviously, the total number of independent parameters is $N^{O_d} \times (N-1)$, which grows exponentially with the order $O_d$, so that it's impossible to achieve effective parameters estimation perfectly.

\subsection{HMM-based note segmentation}

As a replacement of the simple thresholding (see Sec. \ref{ThreshdolingNoteDec}), HMM can be used to perform pitch-wise note segmentation from a salience matrix. This function of HMMs consists in a time filtering of note detection decision and generates smoothed note boundaries. We propose two HMM models for this task, the first order two-state on/off HMMs, developed by \citep{Poliner2007}, and an original one which extends this model by including note duration modeling through a high-order HMM. Here, the hidden stochastic process for HMMs in AMT is associated to the activation matrix $P(i,t)$ computed during the multi-pitch estimation stage.

\begin{figure}[t]
  \centering
  \centerline{
  \includegraphics[width=\columnwidth]{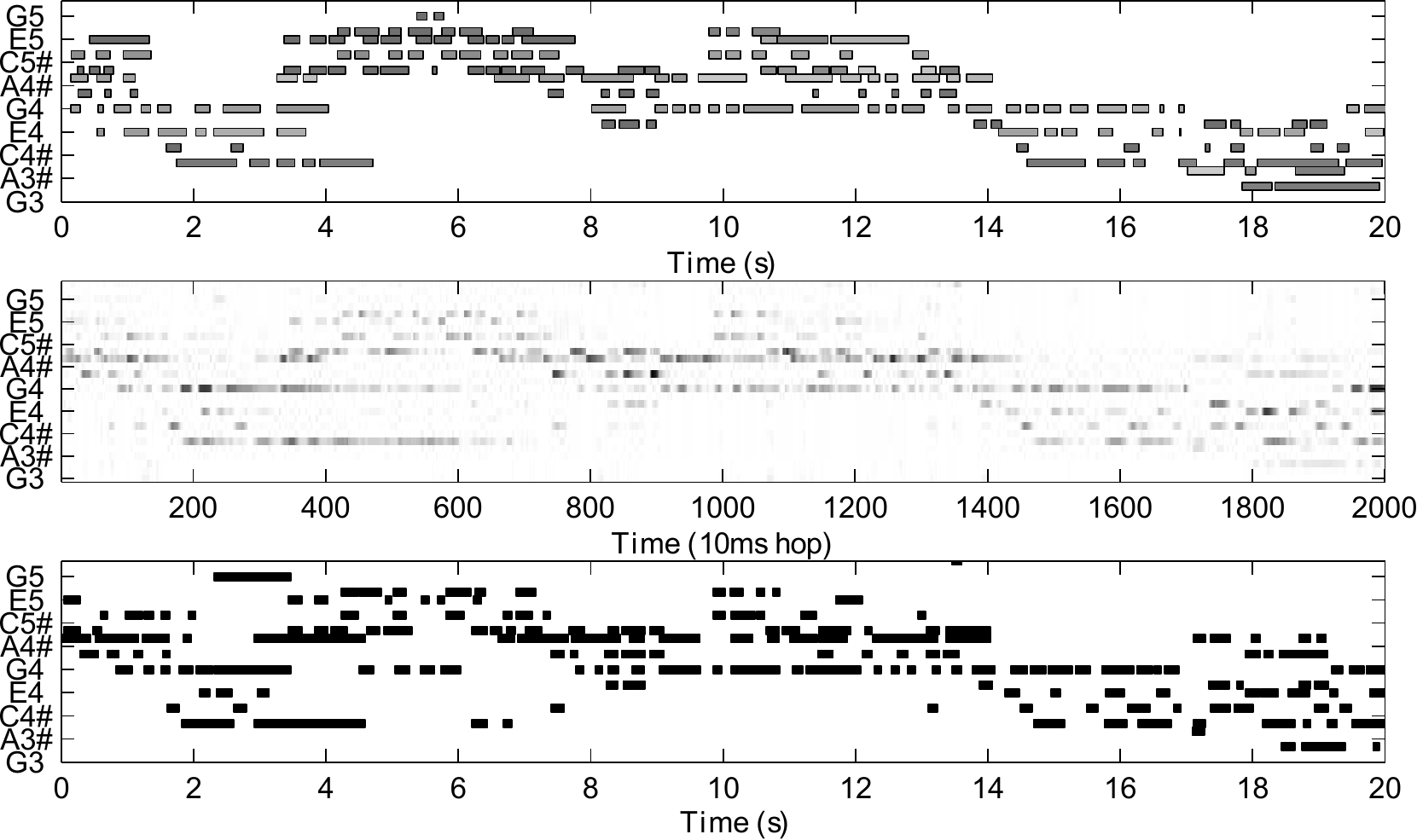}
  }
  \caption{Illustration of different stages of our AMT system on a test musical sequence, with from top to bottom: ground truth, PLCA-based multi-pitch estimation and piano-roll transcription output using system 1.}
  \label{BlockProcessingHMMs}
\end{figure}

\subsubsection{First-order two-state on/off HMM}

This model has been introduced in \citep{Poliner2007}. Each pitch is modelled as a two-state on/off HMM, i.e. $S_i$ $\in$ $\{0,1\}$, which denotes pitch activity/inactivity. The state dynamics, transition matrix, and state priors are estimated from our ``directly observed" state sequences, i.e. the training MIDI transcripts, which are sampled at the precise times corresponding to the analysis frames of the activation matrix. The initial probability $P(q_1)$ for each note is supposed to be 1 for the off state, because all notes are inactive at the beginning of a recording. There are only four transition probabilities $P(q_t | q_{t-1})$, which correspond to the following state transitions: on/on, on/off, off/on, off/off. These probabilities are strongly dependent on the tempo of the considered composition. $P(y_t | q_{t} ) $ is the observation probability, whose values are extracted frame-wisely from the PLCA activation matrix.

\subsubsection{$O_d$-Order two-state on/off HMM}

In this HMM model, as we only have two states, $S_i$ $\in$ $\{0,1\}$, the left-to-right HMM topology allows for a simple modeling of their temporal dynamic through the duration that the last state has stayed. Through this model, we no longer consider consecutive states belonging to a same state as independent identically distributed, but aggregate them in groups according to the duration of staying in a state. The knowledge memorized in this modeling is then about note duration, and allows for an efficient reduction of the HMM parameters to be estimated. Mathematically, eq. \ref{EqHOHMM} can be reduced to

\begin{figure}
\centering
\resizebox{12cm}{!}{
\def\svgwidth{9cm}
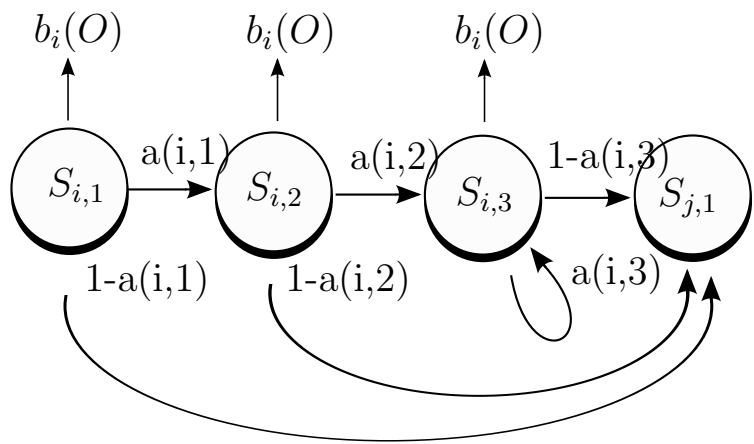
}
\caption{Graphical representation of transition probabilities between three hidden states} 
\label{DOHMM_StateTransition}
\end{figure}

\vspace{-0.5cm}
\begin{multline}
P(q_1,..., q_T,y_1,..., y_T ) = P(q_1)P(y_1 | q_1 , d_1(q_1)) 
\\
\cdot  P(q_t | q_{t-1}, d_{t-1}(q_{t-1})) P(y_t | q_t, d_t(q_t) )
\\
\cdot P(q_T | q_{T-1}, d_{T-1}(q_{T-1})) P(y_T | q_T, d_T(q_T))
\end{multline}

where $d_t(s_t)$ $\in$ $\{1,...,O_d\}$, represents the duration that state $q_t$ has stayed up to time t, and is equal to $O_d$ when it exceeds the maximum dependency order $O_d$. Figure \ref{DOHMM_StateTransition} provides a graphical representation of transition probabilities between three hidden states with this model, along with their observation distribution. We now need to define the score of the candidate optimal partial, which will depend on state i and staying duration d, at a given time t. Such a score can be defined as, for 0 $\leq$ d $\leq$ $O_d$, 

\vspace{-0.5cm}
\begin{multline}
\delta(i,d,t) = \max\limits_{q_{1},...,q_{t-d-1}} ln( P(q_1,..., q_{t-d-1}, \\ 
q_{t-d}=i-1, q_{t-d+1} = \cdots = q_{t}=i , y_1,..., y_T) )
\end{multline}

and for d=$O_d$, 

\vspace{-0.5cm}
\begin{multline}
\delta(i,D,t) = \max\limits_{q_{1},...,q_{t-O_d-1}} ln( P(q_1,..., q_{t-O_d}, \\ 
q_{t-D+1} = \cdots q_{t}= i , y_1,..., y_T ) )
\end{multline}

This HO-HMM model is actually equivalent to $O_d$N-state first order HMM and can then be recursively solved with a Viterbi algorithm. Algorithm \ref{ModViterAlgo} presents this resolution for two on/off states $S_i$ $\in$ $\{0,1\}$. In this algorithm, the probability of transition from state i to itself after staying for d frames is denoted by $\tilde{a}(i,d)$. Briefly, in the initialization step (step \ref{AlgoVitMod_Init}) of the algorithm, the first frame is constrained to be in state $S_1$=0 (i.e. a sequence begins with silence). For d=1 (step \ref{AlgoVitMod_d1}), the first entrance of state $S_1$ is from state $S_2$ with a stay time of 1 through $O_d$, and reciprocally. For d $>$ 2 (step \ref{AlgoVitMod_d2}), the previous frame must be at the same state with stay time equal to d-1. For d =$O_d$ (step \ref{AlgoVitMod_dM}), the first entrance of state $S_1$ is from state $S_2$ with a stay time at state $O_d-1$, and reciprocally. In the termination step (step \ref{AlgoVitMod_Termi}), the last frame is constrained to be at state $S_1$=0, and it should be at the end boundary state after the last frame.

One major advantage of this algorithm is that it allows the complex topology of equivalent first order HMMs to be automatically learned from the training data. Initially, the training data are uniformly segmented according to the state number. The observation vectors belonging to the same state and staying time are grouped. From the uniformly segmented state sequence, we can accumulate the state transition counts and then estimate state transition probabilities. Let C(i,d), d $\leq$ $O_d$, denote the times that state i has stayed d frames and let C(i,$O_d$+) denote the times that state i has stayed for more than $O_d$ frames. Then, the state transition probabilities can be estimated by 

\begin{equation}
\tilde{a}(i,d) = \frac{C(i,d+1)}{C(i,d)}, 1 \leq d<O_d
 \end{equation}
\begin{equation}
\tilde{a}(i,O_d) = \frac{C(i,O_d+)}{C(i,O_d)}
 \end{equation}

\begin{algorithm}
\centering 

\caption{Viterbi Algorithm adapted for a $O_d$-order two-state on/off HMM}\label{ModViterAlgo}
\begin{algorithmic}[1]

\State\label{AlgoVitMod_Init}{Initialization (t=1)} 

\State $\delta(1,1,1) = ln(b_{11}(y_1))$
\State $\delta(1,i,d) = - \inf, \quad \text{for} \quad i \in \{1,2\}, 1<d<O_d$

\For{1$<$t$<$T}

\For{$i,j \in \{1,2\}$, with i $\neq$ j} 

\State\label{AlgoVitMod_d1} $\delta(t,i,1) = \max\limits_{1 < \tau < O_d} \delta(t-1,i-1,\tau)  + ln( \tilde{a}(j,\tau)) + ln(b_{i1}(y_t))$

\For{$2 < d < O_d$} 

\State\label{AlgoVitMod_d2} $\delta(t,i,d) = \delta(t-1,i,d-1) + ln(\tilde{a}(i,d-1)) + ln(b_{id}(y_t))$

\EndFor

\State\label{AlgoVitMod_dM}  $\delta(t,i,O_d) = \max\limits_{O_d-1 < \tau < O_d} \delta(t-1,j,\tau)  + ln(\tilde{a}(i,\tau)) + ln(b_{iO_d}(y_t))$

\EndFor

\EndFor

\State{Termination (t=T) :} 

\State\label{AlgoVitMod_Termi}  $\delta_{opt} = \max\limits_{1 < d < O_d} \delta(T,1,d) + ln( \tilde{a}(2,d) )$

\end{algorithmic}
\end{algorithm}

\subsection{HMM-based post-processing}

This post-processing stage consists in identifying the correct candidate note events among the list \textbf{$L_N$}(i,k) based on polyphonic harmonic transitions. To do so, we will consider these events as the states of $N_c$-state HMMs.

\subsubsection{State generation}\label{StateGene}

In our HMM post-processing, states are not known beforehand, but are defined based on the pitch activation matrix $P(i,t)$. First, instead of keeping a list of binary note candidates \textbf{$L_N$}(i,k), we create probabilistic note candidates \textbf{$\hat{L}_N$}(i,k), on which other post-processing operations can be done. We then need to assign an observation probability likelihood P(\textbf{$\hat{L}_N$}(i,k)) to each note candidate, which has been simply defined as the median of the pitch activation scores contained in the activation interval $T_{i,k}$. In mathematical terms, we have 

\begin{equation}
P(\boldsymbol{\hat{L}_N}(i,k)) = \underset{t \in T_{i,k}}{median}(P(i,t)), \quad \forall i,k
\end{equation}

\begin{figure}[t]
  \centering
  \centerline{
  \includegraphics[width=\columnwidth]{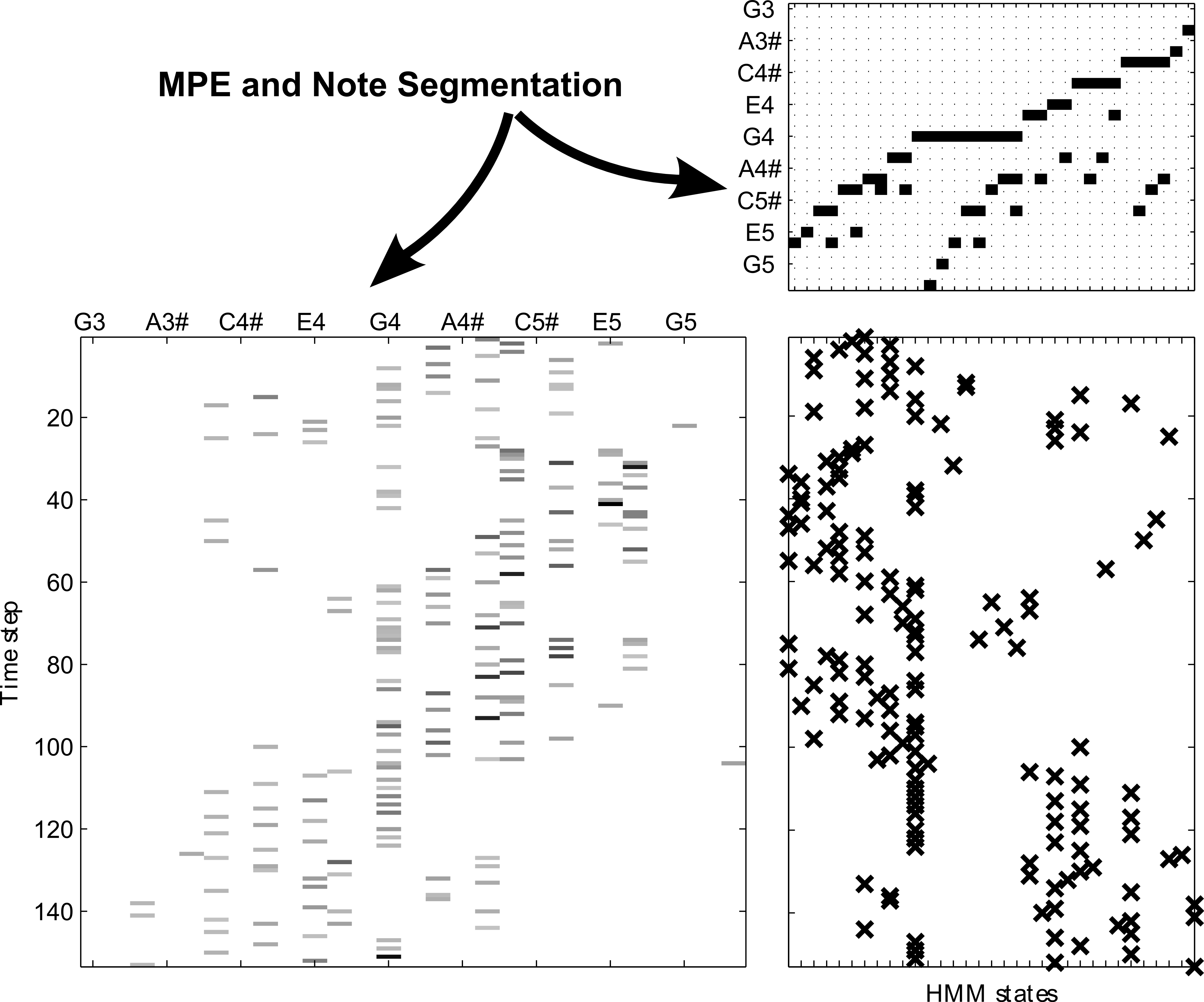}
  }
  \caption{Illustration of HMM state generation, performed after multi-pitch estimation and note segmentation. On the right top graph, the different states are represented vertically with their constitutive pitches. On the graph just below, their temporal distribution is plotted, with the time at the vertical. On the left bottom graph, the observation probabilities of each note candidate $N_c$ are plotted against time. The irregular time step is defined as the successive inter-note intervals.}
  \label{HMM_StateGeneration}
\end{figure}

Then, in order to reduce the number of possible states, we performed the following operations on P(i,t) : 1. pruning with a minimum duration set to 50 ms, followed by 2. an alignment of onsets, i.e. onsets closer than 20 ms are aggregated in a same note mixture, whose onset is computed as the median of note onsets in the mixture, and eventually, 3. equalizing chord duration, i.e. the durations of all notes in a chord are set to a same arbitrary duration. In the following, $N_c$-state HMMs are formed with the different mixtures encountered in the test sequences. It is noteworthy that with the procedure described above, we reduce considerably the number of HMM states, and only keep more salient note mixtures. In the following, we use ergodic models to allow every possible transition from state to state. This overall process of state generation is illustrated in figure \ref{HMM_StateGeneration}.

\subsubsection{First-order $N_c$-state HMM for harmonic transitions}

\begin{figure}

\centering
\resizebox{12cm}{!}{
\def\svgwidth{8cm}
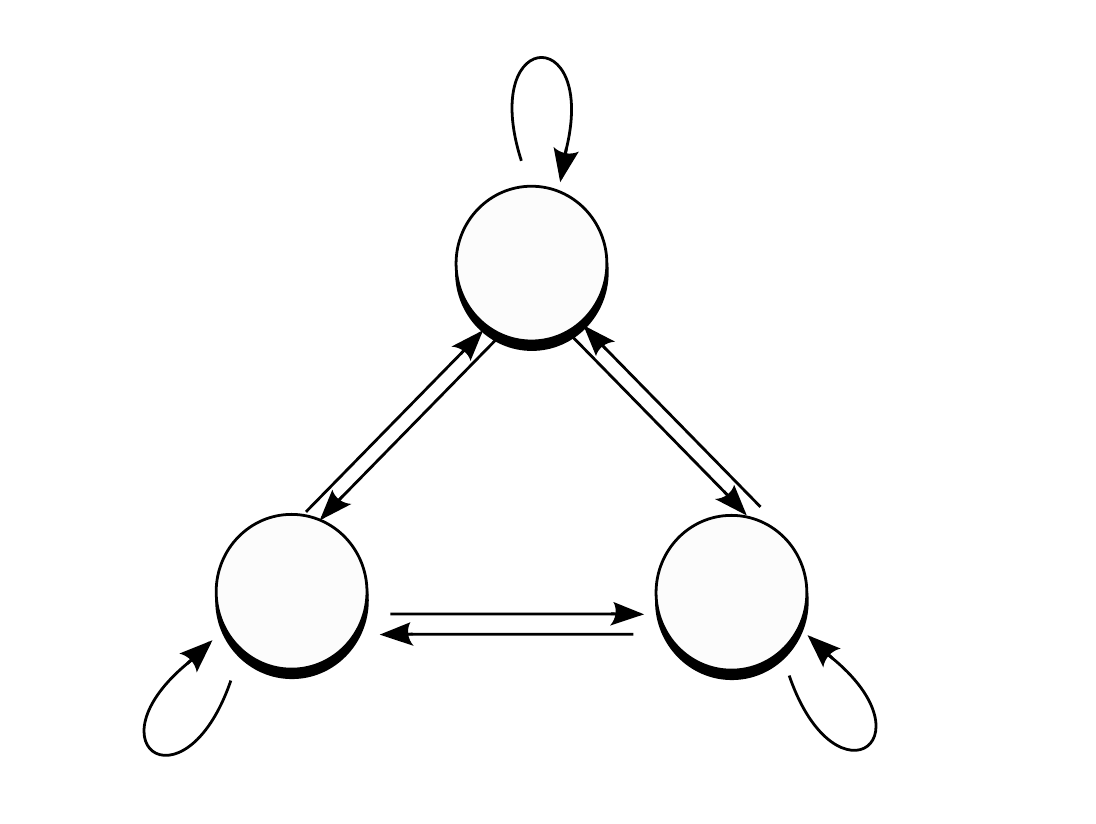
}
\caption{Graphical representation of transition probabilities between three hidden states.} 
\label{FirstOrderStateTrans}
\end{figure}

This model follows a classical first-order HMM with $N_c$ states. Figure \ref{FirstOrderStateTrans} provides a graphical representation of transition probabilities between three hidden states with this model, along with their observation distribution. In this model, transition probabilities are defined as the probability to switch between two successive mixtures of notes in a musical piece, simply computed by counting their occurrence frequency. These probabilities give us a global view of the usual and unusual harmonic transitions of an instrument repertoire, and are trained using MIDI scores. Despite the size of the learning database, novel note transitions may appear in test sequences without being trained, which will then not have estimated likelihoods. The distribution values must be smoothed to address tiny likelihood values for the note transitions that were not found in the learning database. Then, the Witten-Bell discounting algorithm \citep{Witten1991} can be used to perform this smoothing.

\subsubsection{Tonality-based first-order $N_c$-state HMM  for harmonic transitions}

The term tonality (or key) is usually defined as the relationship between a set of pitches having a tonic as its main tone, after which the key is named. A key is then defined by both its tonic and its mode, and generates expectations favouring certain pitch sequences. The tonic is one in an octave range, within the 12 semitones of the chromatic scale. The mode is usually minor or major, depending on the used scale. The major and minor keys then rise to a total set of 24 different tonalities. The prior knowledge of the key results from an automatic detection based on a chroma-based frequency analysis \citep{Shenoy2004}, performed on each training and test sequence. This knowledge is incorporated in a first-order $N_c$-state HMM with the same frequency counting as previously, although now we split the different countings of each sequence according to their respective keys. Thus, we build 24 key-dependent HMMs \citep{Lee2008}, defining a unique state-transition characteristic for each key model.



\subsubsection{Second-order $N_c$-state HMM for harmonic transitions}

We used the extended second-order Viterbi algorithm already published in the literature \citep{He1988}. A same procedure for state generation (see Sec. \ref{StateGene}) has been used. Only now, training of state transition probabilities implies counting two-state sequences.

\begin{table}[h]
\begin{center}
\resizebox{8cm}{!}{
\begin{tabular}{|c|c|c|}
\hline
AMT systems & Note segmentation & Sequential post-processing                                                                              \\ \hline
$M_1$ (baseline)   & Simple thresholding                                                                     & --                                                                                           \\ \hline
$M_2$   & Two-state on/off HMMs                                                                   & --                                                                                           \\ \hline
$M_3$   & \begin{tabular}[c]{@{}c@{}}High-order two-state on/off HMMs \\ for duration modeling\end{tabular} & --                                                                                           \\ \hline
$M_4$   & Simple thresholding                                                                     & \begin{tabular}[c]{@{}c@{}}First-order M-state HMM \\ for harmonic transitions\end{tabular}  \\ \hline
$M_5$   & Simple thresholding                                                                     & \begin{tabular}[c]{@{}c@{}}Second-order M-state HMM \\ for harmonic transitions\end{tabular} \\ \hline
\end{tabular}
}
\caption{Configurations of our different PLCA plus HMM systems for AMT.}\label{MethodsPLCA_HMM}
\end{center}
\end{table}

\subsection{Evaluation procedure}\label{SoundDatabase}

\subsubsection{Musical corpus}

To test our AMT system and train the sparse priors proposed, we need different sound corpus: audio musical pieces of an instrument repertoire, the corresponding scores in the form of MIDI files, as well as a complete dataset of isolated notes for this instrument. We will use two different decay instruments for evaluation, namely the classical piano and the \textit{marovany} zither from Madagascar. For piano, all sound material was extracted from the MAPS database \citep{Emiya2010}, which is composed high-quality note samples and recordings from a real upright piano, whose MIDI scores have been automatically compiled using the Disklavier technology. For the marovany instrument, sound templates and musical pieces were extracted from personal recordings made in our laboratory. Original compositions were transcribed with an original multi-sensor retrieval system \citep{Cazau2013c}. 

From these sound databases, we extracted different sets of training and test data, as our prior must be trained using automatically generated knowledge from MIDI files and template datasets. To do so, we first divided each musical pieces into 30-second sequences, which provided us with a total of 1.2 and 0.83 hours of audio, respectively for the piano and marovany datasets. Within each dataset, the musical sequences were randomly split into training and testing sequences, using by default 30 \% of sequences for testing, and the 70\% remaining ones for training. In our simulation experiments, this procedure is repeated five times, and an average is computed on the resulting scores. To prevent any overfitting of our data, we carefully distinguished between training and test data, i.e. sequences used to train musicological knowledge were not used for evaluation.

\subsubsection{Error metrics}

For assessing the performance of our proposed transcription system, we adopt a note-oriented approach, according to which a note event is assumed to be correct if it fills the condition that its onset is within a 50 ms range from a ground-truth onset (i.e. the standard tolerance commonly used \citep{Mirex2011}). Such a tolerance level is considered to be a fair margin for an accurate transcription, although it is far more tolerant than human ears would, as we remind that those are able to distinguish between two onsets as close as 10 ms apart \citep{Moore1997}. Evaluation metrics are defined by equations \ref{Metr1}-\ref{Metr4} \citep{Mirex2011}, resulting in the note-based recall (TPR), precision (PPV) and F-measure (the harmonic mean of precision and recall) :

\begin{equation}\label{Metr1}
\text{TPR} = \frac{\sum_{n=1}^N {\text{TP}[n]}}{\sum_{n=1}^N {\text{TP}[n] + \text{FN}[n]}} \end{equation}
\begin{equation}\label{Metr2}
\text{PPV} = \frac{\sum_{n=1}^N {\text{TP}[n]}}{\sum_{n=1}^N {\text{TP}[n] + \text{FP}[n]}} 
\end{equation}
\begin{equation}\label{Metr4}
\text{F-measure} = \frac{2 . \text{PPV} . \text{TPR}}{\text{PPV}+\text{TPR}} 
\end{equation}

where N is the total number of notes, and TP, FP and FN scores stand for the well-known True Positive, False Positive and False Negative detections. The recall is the ratio between the number of relevant and original items; the precision is the ratio between the number of relevant and detected items; and the F-measure is the harmonic mean between precision and recall. For all these evaluation metrics, a value of 1 represents a perfect match between the estimated transcription and the reference one.

\section{Results and discussion}\label{sec:resu}

We evaluated comparatively the different PLCA plus HMM systems listed in table \ref{MethodsPLCA_HMM}. Tables \ref{Tab1} and \ref{Tab2} present mean transcription results of our different methods for piano and the marovany recordings, respectively. When comparing PLCA baseline systems (first lines of tables), lower transcription performance has been observed for the marovany instrument repertoire, with respectively mean F-measures of 63.1 and 54.8. From these reference scores, some PLCA plus HMM systems achieved improvements up to + 3.3 $\%$ in the average F-measure values, which are far from negligible when considering current improving results reported in AMT studies. We can also observe that for both repertoires, methods $M_2$ and methods $M_4$ and $M_5$ bring enhancements on methods $M_1$ and $M_3$, respectively.

\begin{table}[h]
\centering
\begin{tabular}{|c|c|c|c|}
\hline
AMT systems           & TPR & PPV & F-measure       \\ \hline
$M_1$ (baseline)          & 62.5 & 63.8 & 63.1  \\ \hline
$M_2$       & 62.7  & 64.1  & 63.4 \\ \hline
$M_3$       & 63.9 & 64.6 & 64.2  \\ \hline
$M_4$       & 65.1 & 66.3 & 65.7  \\ \hline
$M_5$       & 66.1 & 65.7 & 65.9  \\ \hline

\end{tabular}
\caption{Mean transcription error metrics for the piano recordings with our different AMT methods.}\label{Tab1}
\end{table}

The two instrument repertoires used for evaluation in this paper are also differentiated through the transcription performance of the different systems. The piano repertoire benefits most from the tonality-based HMM, as well as the second-order HMM, i.e. systems $M_4$ (with gain of + 2.6 in the F-measure from system $M_1$) and $M_5$ (with gain of + 2.8 in the F-measure from system $M_1$), whereas the marovany repertoire benefits more from the note segmentation performed with the $O_d$ two-state HMM  (with gain of + 3.3 in the F-measure from system $M_1$). Explanations based on the respective characteristics of their repertoires can be put forward here. The marovany repertoire is characterized by fast arpeggios, without a vertical polyphonic writing properly speaking. A rather simple musical polyphony which is compensated by the more complex resonating behavior of the instrument, creating an ample halo-like sound with rich overtones. The two-state HMMs with their smoothing function allow mainly to avoid a lot of single miss errors and short spurious notes, resulting in an accuracy increase in terms of onset detection. Such transcription results are more related to short-term acoustic features (e.g. octave error due to template ambiguity), which fits well to the marovany repertoire. Furthermore, traditional musical playing of the marovany consists in interfering at the minimum on its intrinsic timbre, and even exciting it as louder as possible (an effect called \textit{mafo be}). This characteristic generates robust prior knowledge on note duration with a weak inter-repertoire variability. On the contrary, the classical piano repertoire presents more complex and richer chord transitions, with different playing techniques and dynamics which  interfere continuously on the timbre of the instrument. These characteristics are better captured by a more complex musicological modeling of music structure, in comparison to the marovany repertoire whose harmonic transitions are pretty simple and predictable.

\begin{table}[h]
\centering
\begin{tabular}{|c|c|c|c|}
\hline
AMT systems      & TPR & PPV & F-measure       \\ \hline
$M_1$ (baseline)             & 55.7 & 53.9 & 54.8  \\ \hline
$M_2$       & 59.7 & 53.6 & 56.5  \\ \hline

$M_3$       & 58.7 & 57.6  & 58.1 \\ \hline
$M_4$       & 58.4 & 54.3 & 56.3  \\ \hline

$M_5$       & 56.9 & 57.3 & 57 \\ \hline

\end{tabular}
\caption{Mean transcription error metrics for the marovany recordings with our different AMT methods.}\label{Tab2}
\end{table}

\section{Conclusion}

In this paper, we investigate the use of Hidden-Markov Models in automatic transcription of music. Our results suggest that by optimizing the use of HMM models in AMT systems, valuable enhancements could be brought in transcription performance.

We also would like to convey the idea that the prior characterization of an instrument repertoire, in particular through domain-specific knowledge of other research fields (here, acoustics and musicology), could help integrating more optimal and specific prior knowledge in AMT systems. In future works, a better understanding of the link between HMMs and repertoire characteristics will be investigated.


\section{Acknowledgement}\label{sec:ack}

At the risk of omitting some relevant names, the authors would like to especially thank Marc Chemillier (CAMS-EHESS) for recordings of the \textit{marovany} and Laurent Quartier (LAM-UPMC) for technical supports.

\bibliography{References_Biblio}
\bibliographystyle{Cazau}

\end{document}